\documentclass{article}

\usepackage[final]{nips_2018}

\usepackage[utf8]{inputenc} 
\usepackage[T1]{fontenc}    
\usepackage{hyperref}       
\usepackage{url}            
\usepackage{booktabs}       
\usepackage{amsfonts}       
\usepackage{nicefrac}       
\usepackage{microtype}      
\usepackage{graphicx}
\usepackage{comment}
\usepackage{amsmath}

\title{Practical Window Setting Optimization \\ for Medical Image Deep Learning}

\author{
    Hyunkwang Lee \\
    School of Engineering and Applied Sciences\\
    Harvard University\\
    Cambridge, MA 02138 \\
    \texttt{hyunkwanglee@seas.harvard.edu} \\
    \And 
    Myeongchan Kim\\
    Department of Radiology \\
    Massachusetts General Hospital\\
    Boston, MA 02114 \\
    \texttt{mkim49@mgh.harvard.edu} \\
    \And 
    Synho Do \\
    Department of Radiology \\
    Massachusetts General Hospital\\
    Boston, MA 02114 \\
    \texttt{sdo@mgh.harvard.edu} \\
}

\begin{document}

\maketitle

\begin{abstract}
The recent advancements in deep learning have allowed for numerous applications in computed tomography (CT), with potential to improve diagnostic accuracy, speed of interpretation, and clinical efficiency. However, the deep learning community has to date neglected window display settings -- a key feature of clinical CT interpretation and opportunity for additional optimization. Here we propose a window setting optimization (WSO) module that is fully trainable with convolutional neural networks (CNNs) to find optimal window settings for clinical performance. Our approach was inspired by the method commonly used by practicing radiologists to interpret CT images by adjusting window settings to increase the visualization of certain pathologies. Our approach provides optimal window ranges to enhance the conspicuity of abnormalities, and was used to enable performance enhancement for intracranial hemorrhage and urinary stone detection. On each task, the WSO model outperformed models trained over the full range of Hounsfield unit values in CT images, as well as images windowed with pre-defined settings. The WSO module can be readily applied to any analysis of CT images, and can be further generalized to tasks on other medical imaging modalities. Our codes for the full implementation of window setting optimization module (based on Keras) are available at \url{https://github.com/Synho/windows_optimization.git}.
\end{abstract}
\section{Introduction}
Deep learning has made remarkable advancements in medical image analysis for various tasks across a range of imaging modalities \citep{esteva2017dermatologist, gulshan2016development, chilamkurthy2018deep}. This rapid progress in image analysis capability has raised hopes that deployment of such technology will increase diagnostic accuracy, streamline clinical workflows, and improve patient outcomes \citep{thrall2018artificial, levin2018machine, berlyand2018artificial}. Much of this progress has been attributed to increased computing power and the development of large and well-curated clinical datasets \citep{Yu2018}. However, recent years have demonstrated that significant performance gains remain to be had from innovations in neural network architectures and domain-specific image pre-processing \citep{greenspan2016guest}. 

Deep learning architectures in the medical domain may yet see significant performance improvements by incorporating expert knowledge about the target imaging modality and the current clinical workflow. In the case of computed tomography (CT), image values are defined over a wide range of Housefield Units (HU), but with different tissue types and pathologies generally visible only in narrow and specific ranges. As such, when interpreting CT images, human experts leverage tools in their workstations to apply predefined window levels (WL) and window widths (WW) to their display windows; these window adjustments focus visibility on the subset of tissues relevant to their task, and are crucial for the effective detection of some pathologies \citep{bae2005ct, moise2004design}. For example, radiologists may use a pre-set "brain" or "subdural" window setting for intracranial hemorrhage(ICH) detection. Despite the importance of optimal window settings in clinical practice, however, the effects of window settings on image quality and algorithmic performance have been overlooked in the literature. Most previous works have converted CT images to grayscale with a pre-set window setting, encoded three different windowed images into an RGB image, or used a wide range of image intensity values without windowing as input into deep learning models \citep{arbabshirani2018advanced, hoo2016deep, anthimopoulos2016lung, chang2018hybrid}.

In this study, inspired by the way radiologists interpret CT images, we propose a window setting optimization (WSO) module comprised of convolution layers with 1x1 filters and customized activation functions. This enables us to find optimal window settings in a task-specific manner via backpropogation, which we demonstrate results in improved model performance on the detection of ICH and urinary stones. This WSO module mimics the radiologist's workflow to optimize windowing functions and focus on the narrow window range in which the target organs or abnormalities can be clearly seen. Our method can be potentially applied to other CT image analysis tasks such as object detection and semantic segmentation, or to other medical imaging modalities such as a positron emission tomography (PET) scan and magnetic resonance imaging (MRI).

\section{Methods}
\begin{figure}
\begin{center}
        \includegraphics[scale=0.8]{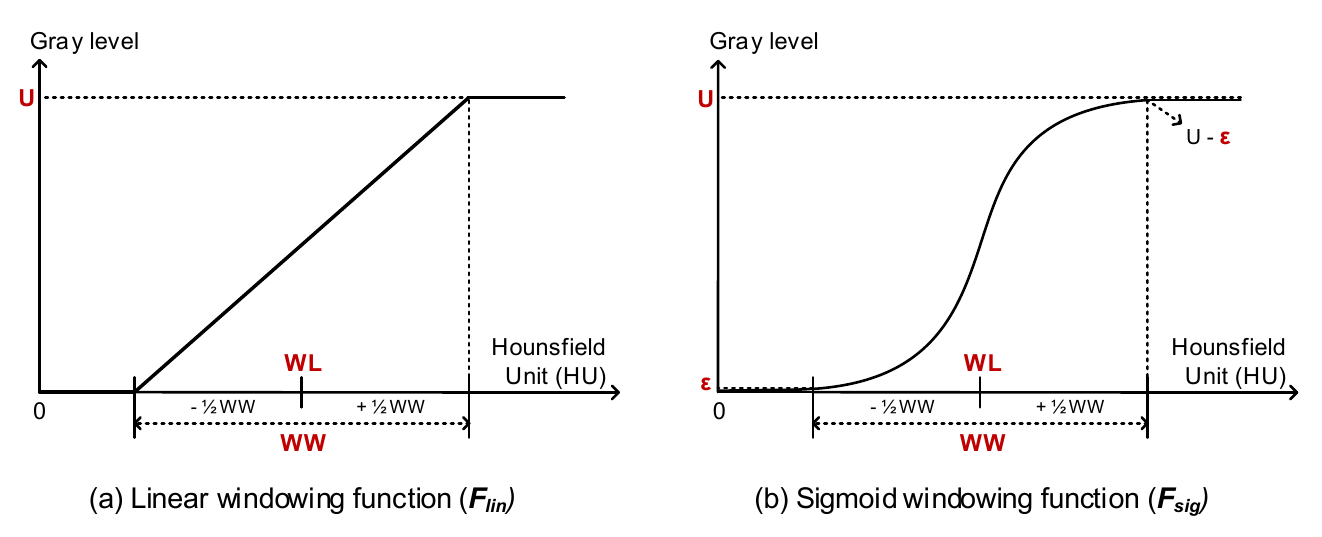}
        \caption{\label{fig:windowfunc}Window functions}
\end{center}
\end{figure}
\subsection{Windowing function}
The most common format for medical images is the digital imaging and communications in medicine (DICOM) format. DICOM images are encoded with either 12 or 16 bits per pixel, which is 4,096 or 65,536 levels per pixel, respectively. In CT imaging, these pixels represent Housefield Units values that correspond to tissue density and are generally distributed from $HU_{air}$(-1000) to >4000. The range and granularity of the data encoded in CT images extends far beyond the perceptual capacity of the human visual system, which can distinguish only several hundreds of grayscale. In addition, most medical displays support at most 8-bit resolution \citep{kimpe2007increasing}. For these technical and biological reasons, CT images can only be successfully interpreted by humans when the display device allows for a window function with adjustment of window settings. These settings map the visual range of the displays to a specified window, and assign all HU values outside this window range to $0$ or $U$ (Fig. \ref{fig:windowfunc}). Windowing functions are defined based on linear or sigmoidal conversion as the following equations:
\begin{equation}\label{eq:linear}
\begin{aligned}
F_{lin}(x) = \min(\max(Wx+b, U), 0),
where~ W=\frac{U}{WW},
~ b=-\frac{U}{WW}(WL-\frac{WW}{2})
\end{aligned}
\end{equation}

\begin{equation}\label{eq:sigmoid}
\begin{aligned}
F_{sig}(x) = \frac{U}{1+e^{-(Wx+b)}},
where~ W=\frac{2}{WW}\log(\frac{U}{\epsilon} - 1),
~ b=\frac{-2WL}{WW}\log(\frac{U}{\epsilon} - 1)
\end{aligned}
\end{equation}
The constant $U$ is the upper limit of windowing functions and $\epsilon$ is the margin between the upper/lower limits and window end/start gray levels which determine the slope at the center.

\subsection{Window setting optimization module}
\begin{figure}
\begin{center}
        \includegraphics[scale=1.0]{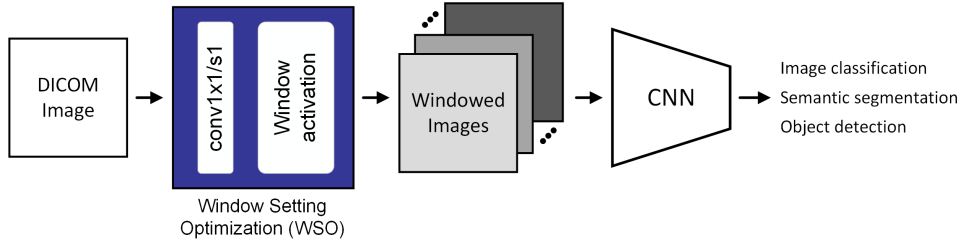}
        \caption{\label{fig:system}Overall architecture of window setting optimization (WSO) module}
\end{center}
\end{figure}

The linear and sigmoid windowing functions utilized in radiologist's monitors can be emulated as a WSO module inside a neural network architecture. This is achieved by using convolution layers with 1x1 filters and a stride setting, followed by an activation layer -- an upper-bounded rectified linear unit (ReLU), or sigmoid function multiplied by $U$, respectively, for the linear or sigmoid windowing function. In our implementation, full-range DICOM images are passed through this WSO module prior to being used as input to a CNN, as shown in Fig. \ref{fig:system}. Weights and biases of the 1x1 convolution layers in the WSO module can thus be optimized along with the CNN, facilitating the identification of optimal windowing functions to visually extract the necessary features for maximum classification performance.

\subsection{Dataset}
This retrospective health insurance portability and accountability act (HIPAA)-compliant study was approved by the Institutional Review Board at our institution. All DICOM images were de-identified prior to this study. For ICH detection, a total of 904 non-contrast head CT examinations, including 625 ICH-positive and 279 ICH-negative cases, were acquired from our institutional Picture Archiving and Communication System (PACS) between June 2003 and July 2017. For urinary stone detection, we retrieved a total of 515 unenhanced abdominopelvic CT examinations, including 256 stone-negative and 279 stone-positive cases, from our PACS between January and October 2016. 2-dimensional (2D) axial slices of all head CT cases were annotated as the presence of ICH by five neuroradiologists by consensus, and axial slices from all abdominopelvic CT scans were labeled by a radiologist with 6 years experience as presence of urinary stones along with original radiology reports. 2D axial slices for both ICH and urinary stone detection were randomly split into train, validation, and test datasets by cases to make sure there is no overlap in cases between subsets (Table \ref{table:data}).

\subsection{Experimental setup}
 We evaluated ten different classification models developed by training Inception-v3 \citep{szegedy2016rethinking} on various forms of input images with and without using WSO for ICH and urinary stone detection. First, CT images with a full dynamic range of HU values were used for a baseline model. Images converted with either of two pre-defined window settings and two-channel images generated with both were also used as input to CNN without a WSO module. The two pre-set window settings include "brain" (WL=50HU, WW=100HU) and "subdural" (WL=50HU, WW=130HU) windows for ICH detection, and "bone" (WL=300HU, WW=1500HU) and "abdomen" (WL=40HU, WW=400HU) for urinary stone detection. In addition, Inception-v3 equipped with WSO was trained on full-range DICOM images with weights and biases of convolution layers in the WSO initialized according to the windowing function type. For this preliminary study, we set $U$ as 255 and $\epsilon$ as 1, and associated $W$ and $b$ were then computed using Eq. \ref{eq:linear} for linear and Eq. \ref{eq:sigmoid} for sigmoid for each pre-defined window setting. All classification models were trained for 60 epochs using the Adam optimizer \citep{kingma2014adam} with default settings and a mini-batch size of 64. The base learning rate of 0.001 was decayed by a factor of 10 every 20 epochs, and the best models were selected based on the validation loss.
\section{Results}
\begin{table}[!htbp]
\centering
\scalebox{0.9}{
\begin{tabular}{c|cc|cc|cc}
    \toprule
     & \multicolumn{2}{c|}{WSO} & 
    \multicolumn{2}{c|}{ICH} & 
    \multicolumn{2}{c}{Stone}\\
    Input
    & \multicolumn{1}{c}{Windowing function} & \multicolumn{1}{c|}{Initialization}
    & AP & AUC & AP & AUC\\
    \hline
    \midrule
    \multicolumn{1}{r}{HU values} 
    & \multicolumn{1}{c}{-}&\multicolumn{1}{c|}{-}&0.807&0.923&0.813&0.800\\
    
    \multicolumn{1}{r}{Windowed with $S_1$}
    & \multicolumn{1}{c}{-}&\multicolumn{1}{c|}{-}&0.925&0.963&0.920&0.917\\
    
    \multicolumn{1}{r}{Windowed with $S_2$}
    & \multicolumn{1}{c}{-}&\multicolumn{1}{c|}{-}&0.932&0.967&0.945&0.944\\
    
    \multicolumn{1}{r}{Windowed with $S_1$, $S_2$}
    & \multicolumn{1}{c}{-}&\multicolumn{1}{c|}{-}&0.934&0.969&0.946&0.946\\    
    
    \multicolumn{1}{r}{HU values}
    & \multicolumn{1}{c}{$F_{lin}$}
    & \multicolumn{1}{c|}{$S_1$}
    &0.929&0.963&0.926&0.924\\ 
    
    \multicolumn{1}{r}{HU values}
    & \multicolumn{1}{c}{$F_{lin}$}
    & \multicolumn{1}{c|}{$S_2$}
    & 0.933&0.966&0.943&0.934\\ 
    
    \multicolumn{1}{r}{HU values}
    & \multicolumn{1}{c}{$F_{lin}$}
    & \multicolumn{1}{c|}{$S_1$, $S_2$} 
    & 0.940&0.970&0.951&0.946\\ 
    
    \multicolumn{1}{r}{HU values}
    & \multicolumn{1}{c}{$F_{sig}$}
    & \multicolumn{1}{c|}{$S_1$}
    & 0.930&0.966&0.959&0.955\\ 
    
    \multicolumn{1}{r}{HU values}
    & \multicolumn{1}{c}{$F_{sig}$} 
    & \multicolumn{1}{c|}{$S_2$}
    & 0.939 & 0.971 & 0.970 & 0.970\\ 
    
    \multicolumn{1}{r}{HU values}
    & \multicolumn{1}{c}{$F_{sig}$}
    & \multicolumn{1}{c|}{$S_1$, $S_2$}
    & \textbf{0.950} & \textbf{0.976} & \textbf{0.971} & \textbf{0.972}\\ 
    \bottomrule
\end{tabular}}
\caption{\label{table:results} Performance of ten different models trained with different input data and WSO module for ICH and urinary stone detection. $S_1$=Brain window and $S_2$=Subdural window settings for ICH and $S_1$=Bone window and $S_2$=Abdomen window settings for Stone.}
\end{table}
Table \ref{table:results} shows average precision (AP) and area under the ROC curve (AUC) evaluated on the test sets for the ten different models for ICH and urinary stone detection. Models trained on windowed images with pre-defined settings obtained significantly better performance than models trained with CT images with a dynamic full range of HU values for both classification tasks. Furthermore, model performance was improved when using WSO to optimize window settings instead of using fixed standard ones, especially with sigmoid windowing function, parameters of which were initialized according to two standard window settings.
\begin{figure}
\begin{center}
        \includegraphics[scale=0.75]{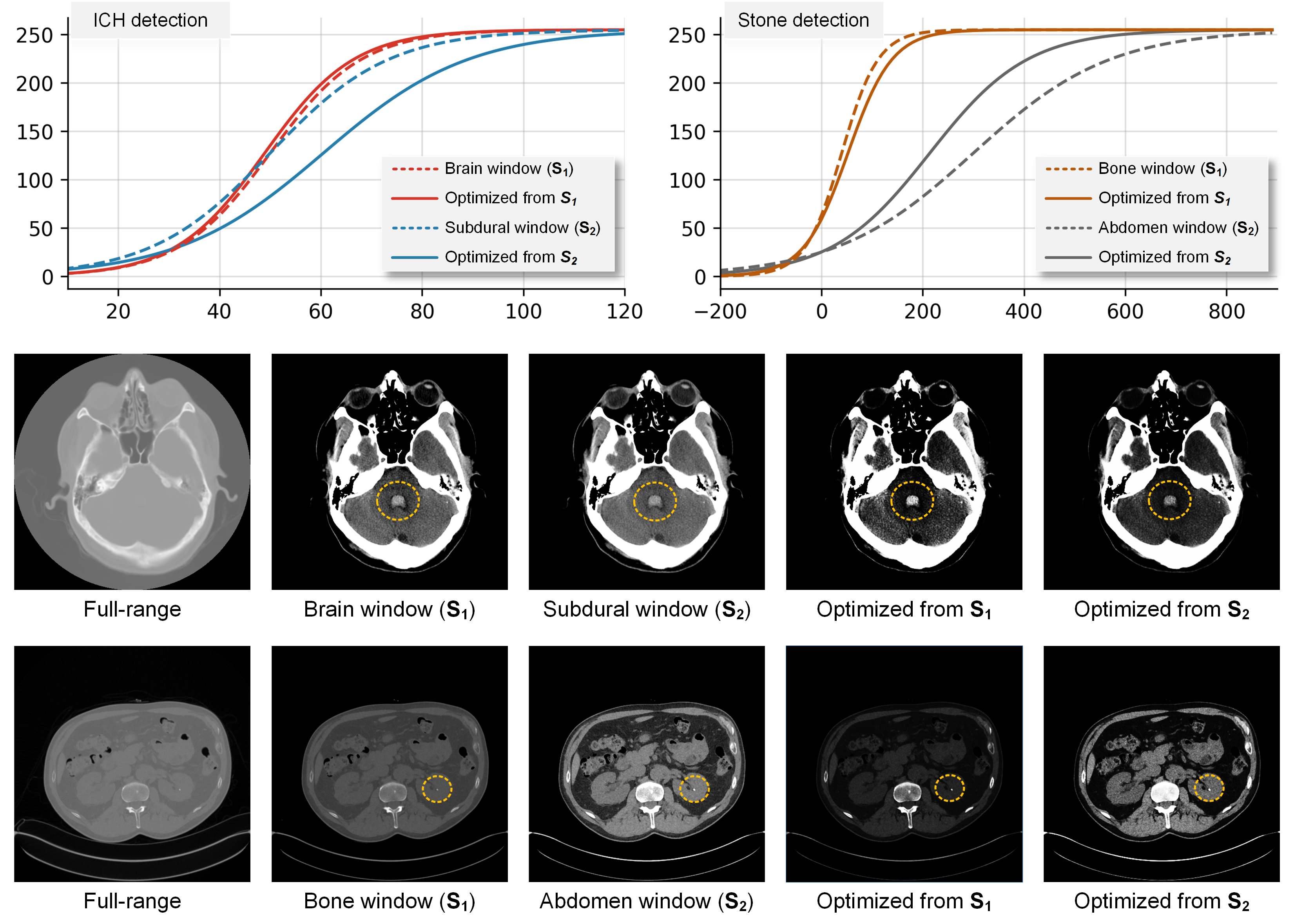}
        \caption{\label{fig:windowimages}Examples of DICOM and optimized windowed images}
\end{center}
\end{figure}
\section{Discussion}
In this study, we demonstrated that models with WSO achieved better performance for ICH and urinary stone detection on CT images, compared to using full-range DICOM images or windowed images with standard pre-defined window settings. Furthermore, as shown in Figure \ref{fig:windowimages}, WSO enabled models to find optimal window settings that make the regions of hemorrhage and urinary stones (highlighted in yellow) more conspicuous against neighboring anatomical structures to maximize classification model performance. Our WSO models can be further optimized by investigating the effects of the number of input image channels, $\epsilon$, and $U$ on the performance of target application. Additionally, we stress that the WSO-based approach described here is not specific to abnormality classification on CT images, but rather generalizable to various image interpretation task on a variety of medical imaging modalities.

\section{Acknowledgements}
We thank Claire Jeon and Samuel G. Finlayson for proofreading and constructive comments on the manuscript.

\bibliographystyle{ACM-Reference-Format}
\bibliography{main}

\newpage

\section*{Appendix}

\begin{table}[ht]
\centering
\scalebox{0.8}{
\begin{tabular}{c|cc|cc|cc|cc}
    \toprule
    \multicolumn{1}{l|}{} &
    \multicolumn{4}{c|}{ICH} & 
    \multicolumn{4}{c}{Stone} \\
    \multicolumn{1}{l|}{} & 
    \multicolumn{2}{c}{No ICH} & \multicolumn{2}{c|}{ICH} &
    \multicolumn{2}{c}{No Stone} & \multicolumn{2}{c}{Stone} \\
    \multicolumn{1}{l|}{Data split} &
    \multicolumn{1}{c}{no. cases} & \multicolumn{1}{c}{no. slices} &
    \multicolumn{1}{c}{no. cases} & \multicolumn{1}{c|}{no. slices} &
    \multicolumn{1}{c}{no. cases} & \multicolumn{1}{c}{no. slices} &
    \multicolumn{1}{c}{no. cases} & \multicolumn{1}{c}{no. slices} \\
    \midrule
    \multicolumn{1}{r|}{Train} & 179 & 7484 & 525 & 5517 & 176 & 1179 & 199 & 1179 \\
    \multicolumn{1}{r|}{Validation} & 50 & 2185 & 50 & 668 & 30 & 181 & 30 & 181 \\
    \multicolumn{1}{r|}{Test} & 50 & 2139 & 50 & 613 & 50 & 347 & 50 & 347 \\
    \bottomrule
\end{tabular}
}
\caption{\label{table:data}Data distribution for ICH and urinary stone detection tasks}
\end{table}

\end{document}